\PassOptionsToPackage{prologue,dvipsnames}{xcolor}
\documentclass[10pt, a4paper]{article}
\usepackage{lrec-coling2024} 
\usepackage[utf8]{inputenc} 
\usepackage[T1]{fontenc}   
\usepackage{colortbl}
\definecolor{color1}{RGB}{0,100,0}
\definecolor{color2}{RGB}{227,38,54}
\usepackage{booktabs} 
\usepackage{multirow}
\usepackage{color} 
\usepackage{diagbox}
\usepackage{tabularx}
\usepackage{adjustbox}
\usepackage{float}
\usepackage{threeparttable}
\usepackage{wrapfig}
\usepackage{CJK}
\usepackage{array}
\usepackage{caption}        
\usepackage{subcaption}
\usepackage{makecell}
\usepackage{rotating}
\usepackage{soul}
\usepackage{comment}
\usepackage{pifont}
\usepackage{enumitem,amssymb}
\usepackage{tcolorbox}
\usepackage{csquotes}
\usepackage{fontawesome}
\usepackage{siunitx}
\usepackage{amsmath}
\newcommand{\cmark}{\textcolor{RoyalBlue}{\ding{51}}}

\title{Can multiple-choice questions really be useful\\in detecting the abilities of LLMs?}
 
\name{Wangyue Li$^{\dagger\diamondsuit}$, Liangzhi Li$^{\dagger\blacklozenge}$\sthanks{\quad Corresponding author.}\hspace{0.15cm}, Tong Xiang$^{\dagger\blacklozenge}$, Xiao Liu$^{\dagger}$, {\bf \large Wei Deng$^{\diamondsuit\heartsuit}$, Noa Garcia$^{\blacklozenge}$}} 

\address{$^{\dagger}$Meetyou AI Lab, $^{\diamondsuit}$Southwestern University of Finance and Economics, $^{\blacklozenge}$Osaka University \\ $^{\heartsuit}$Chongqing University of Posts and Telecommunications \\
\{liliangzhi,liuxiao\}@xiaoyouzi.com\\
\{alee90792,tongxiang39\}@gmail.com\\
dengwei@swufe.edu.cn, noagarcia@ids.osaka-u.ac.jp}

\abstract{
Multiple-choice questions (MCQs) are widely used in the evaluation of large language models (LLMs) due to their simplicity and efficiency. However, there are concerns about whether MCQs can truly measure LLM's capabilities, particularly in knowledge-intensive scenarios where long-form generation (LFG) answers are required. The misalignment between the task and the evaluation method demands a thoughtful analysis of MCQ's efficacy, which we undertake in this paper by evaluating nine LLMs on four question-answering (QA) datasets in two languages: Chinese and English. We identify a significant issue: LLMs exhibit an order sensitivity in bilingual MCQs, favoring answers located at specific positions, i.e., the first position. We further quantify the gap between MCQs and long-form generation questions (LFGQs) by comparing their direct outputs, token logits, and embeddings. Our results reveal a relatively low correlation between answers from MCQs and LFGQs for identical questions. Additionally, we propose two methods to quantify the consistency and confidence of LLMs' output, which can be generalized to other QA evaluation benchmarks. Notably, our analysis challenges the idea that the higher the consistency, the greater the accuracy. We also find MCQs to be less reliable than LFGQs in terms of expected calibration error. Finally, the misalignment between MCQs and LFGQs is not only reflected in the evaluation performance but also in the embedding space. Our code and models can be accessed at~\url{https://github.com/Meetyou-AI-Lab/Can-MC-Evaluate-LLMs}.
 \\ \newline }
\begin{document}

\maketitleabstract

\section{Introduction}

Over the past few years, large language models (LLMs) have exhibited remarkable performance on a wide range of question-answering (QA) tasks~\citep{DBLP:conf/nips/BrownMRSKDNSSAA20,DBLP:journals/corr/abs-2207-05221,DBLP:conf/iclr/RobinsonW23}. The evaluation of LLMs' strengths and limitations often relies on diverse benchmarks presented in different formats~\citep{DBLP:journals/corr/abs-2305-09617,DBLP:journals/corr/abs-2308-07847}, domains~\citep{DBLP:conf/emnlp/JinDLCL19, DBLP:conf/aaai/ZhongXTZ0S20}, and languages~\citep{DBLP:conf/emnlp/PetroniRRLBWM19,DBLP:journals/corr/abs-2302-04023}. As previous research has shown~\citep{DBLP:journals/corr/abs-2211-09110,DBLP:journals/corr/abs-2307-03109,DBLP:journals/corr/abs-2306-09212,DBLP:journals/corr/abs-2306-04757}, evaluation using benchmarks is essential for the detection and mitigation of various issues such as misinformation~\citep{DBLP:conf/icail/ZhengGA0H21, DBLP:journals/corr/abs-2202-13529}, hate speech~\citep{DBLP:conf/emnlp/ElSheriefZMASCY21, DBLP:conf/acl/LuXZMYL23}, and malicious uses~\citep{DBLP:conf/naacl/XuJLBWD21,DBLP:journals/corr/abs-2209-07858, DBLP:conf/acl/Shaikh0HBY23,DBLP:journals/corr/abs-2307-15043}. Such mechanisms are critical for safeguarding against harmful content and promoting responsible usage of LLMs in various contexts.

QA benchmarks come in a variety of formats, including \textit{True/False questions} (TFQs) in which models predict whether a statement in the question is correct or not, \textit{multiple-choice questions} (MCQs), in which multiple candidate answers accompany the input question, and \textit{long-form generation questions} (LFGQs), in which a generated answer could span multiple sentences.
Among these, multiple-choice is the most popular format as it allows a simple and quick assessment of model performance~\citep{DBLP:journals/corr/abs-2102-03315,DBLP:conf/icpr/RamamurthyA22,DBLP:journals/corr/abs-2305-10263,DBLP:journals/corr/abs-2305-08322}. However, MCQs also present several limitations, such as potential misalignment with real-world use cases where LLMs are often required to answer questions in long-from generation format~\citep{Nuance-chatbot,DBLP:journals/corr/abs-2108-07258}.
In addition, LLMs have been shown to be affected by changes in the position of the candidate answers~\citep{DBLP:journals/corr/abs-2306-05685,DBLP:journals/corr/abs-2305-17926} and their contents~\citep{DBLP:journals/corr/abs-2308-11483} when answering MCQs. The aforementioned problems highlight the limitations of MCQs benchmarks in evaluating LLMs, which could potentially lead to overestimation of LLMs capabilities. 

With the above issues in mind, our motivation is to explore the limitations and characteristics of both MCQs and LFGQs as main evaluation formats in QA tasks. We aim to answer the following research questions: 

\begin{enumerate}
    \item How does the arrangement of options in MCQs influence LLMs' selection of responses?
    \item What methodologies can be employed to conduct comprehensive comparative experiments between MCQs and LFGQs? Additionally, what specific aspects should be considered when conducting comparative tests?
\end{enumerate}

The answers to these questions contribute to the understanding and comparison between MCQs and LFGQs as evaluation formats in QA tasks. Given the prevalence of MCQs as the dominant evaluation format, our aim is to thoroughly examine their efficacy. This begins with a detailed exploration of MCQs' capabilities and subsequently extends to a comparative analysis with LFGQs, providing a comprehensive assessment of both formats. 

We address the first question by conducting a series of experiments (\S \ref{Preference}) to reveal the sensitivity of LLMs to answering MCQs by applying slight perturbations to the positional order of the options. We find significant differences between the answers in multiple LLMs (\S \ref{st}). We also identify specific patterns of the selected answer according to its position that varies among different LLMs (\S \ref{pe}). For the second question, we conduct comparative experiments (\S \ref{cp}) to quantify the misalignment between MCQs and LFGQs on three different spaces: direct output space (\S \ref{do}), token logits space (\S \ref{tp}), and hidden embedding space (\S \ref{e}). By doing this, we aim to gain a deeper understanding of the unique characteristics between the two types of questions. 

Our key findings reveal that:

\begin{itemize}[noitemsep,topsep=0pt]
\item LLMs exhibit order sensitivity in bilingual MCQs, favoring answers at the first position.
\item Answers obtained from MCQs and LFGQs for identical questions have a low correlation.
\item Higher consistency does not indicate better model performance.
\item The misalignment between MCQs and LFGQs is evident in the evaluation performance as well as in the embedding space.
\end{itemize}

Overall, our study aims to provide a better understanding of the difference in QA formats in LLM evaluation, uncover underlying patterns, and shed light on the improvement of current methods.

\section{Experimental Details}

\paragraph{Models} 
\label{model}
We use different models on different experiments, tailoring our choices based on the specific goals of each experiment, as summarized in Table \ref{tab:1}.  
To check whether LLMs are sensitive to the order of the candidate answers (\S \ref{Preference}), we evaluate three models: \textbf{ChatGLM-6B}~\citep{DBLP:conf/iclr/ZengLDWL0YXZXTM23,DBLP:conf/acl/DuQLDQY022} and two models from the GPT family, namely \textbf{GPT-3.5-turbo}~\citep{Eco:1990} and \textbf{GPT-4}~\citep{DBLP:journals/corr/abs-2303-08774}.
In comparing MCQs and LFGQs (\S \ref{cp}), we again use different models on the three different spaces. For the direct output space (\S \ref{do}), we use GPT-3.5-turbo, GPT-4, and ChatGLM-6B, considering both their diversity and performances. For the token logits space (\S \ref{tp}), we only test GPT-3.5-turbo, as it is the only model that can output token probabilities~\citep{DBLP:conf/emnlp/ManakulLG23} within three models. Finally, in the embedding space (\S \ref{e}), we conduct experiments with models from multiple popular LLM families across various sizes, including \textbf{StableLM-Tuned-Alpha-3/7B}~\citep{S}, \textbf{RedPajama-INCITE-Instruct-3B-v1}~\citep{RedPajama}, \textbf{Llama-2-7b-chat-hf}~\citep{DBLP:journals/corr/abs-2307-09288}, \textbf{Dolly-v2-2/7/12B}~\citep{DatabricksBlog2023DollyV2}, \textbf{Vicuna-7b-v1.3}~\citep{chiang2023vicuna}, and \textbf{Open-llama-3/7B}~\citep{DBLP:journals/corr/abs-2302-13971}. 

\begin{table*}[t]
\centering
\resizebox{0.98\textwidth}{!}{
\begin{tabular}{lcccccc}
\toprule
 & \multicolumn{2}{c}{\textbf{Are LLMs sensitive to order?}} & & \multicolumn{3}{c}{\textbf{MCQ vs LFGQ}}\\
  \cline{2-3}
  \cline{5-7}
\textbf{Model} & \textbf{Order Sensitivity}  & \textbf{Patterns Decomposition} & & \textbf{Direct Output} & \textbf{Token Logits} & \textbf{Embeddings} \\
\midrule
GPT-3.5-turbo~~\citep{Eco:1990}       & \cmark  & \cmark & & \cmark  & \cmark &\\
GPT-4~\citep{DBLP:journals/corr/abs-2303-08774}                & \cmark & \cmark & & \cmark  &   &  \\
ChatGLM-6B~\citep{DBLP:conf/iclr/ZengLDWL0YXZXTM23}         & \cmark  & \cmark & & \cmark & &\\
Stablelm-tuned-$\alpha$~\citep{S} &  & & & & & \cmark \\
RedPajama-INCITE-3B-v1~\citep{RedPajama}           &  & & & & & \cmark\\
Dolly-v2~\citep{DatabricksBlog2023DollyV2}               &  &  & &  & &\cmark\\ 
Vicuna-7b-v1.3~\citep{chiang2023vicuna}              &  & & & & &\cmark\\
Open-llama~\citep{DBLP:journals/corr/abs-2302-13971}          &  & & & & &\cmark\\
 Llama-2-7b-chat-hf~\citep{DBLP:journals/corr/abs-2307-09288}          &  & & & & &\cmark\\
\bottomrule
\end{tabular}}
\caption{Summary of the LLMs used in each of our analyses.}
\label{tab:1}
\end{table*}

\paragraph{Datasets}
We conduct experiments on four evaluation benchmarks: 

\begin{enumerate}
\item \textbf{CARE-MI}~\citep{CARE-MI}: A Chinese benchmark for evaluating LLM misinformation in the maternity and infant care domain. It includes $1,612$ LFGQs. The questions can also be obtained as MCQs and TFQs from the original MLEC-QA\citep{DBLP:conf/emnlp/LiZC21a} and MEDQA~\citep{DBLP:journals/corr/abs-2009-13081} datasets, according to the question generation process of CARE-MI, resulting in each question being formulated in the three formats: MCQ, LFGQ, and TFQ. 
\item \textbf{M3KE}~\citep{DBLP:journals/corr/abs-2305-10263}:  A dataset with $20,477$ standard Chinese questions for $71$ tasks, encompassing all major levels of Chinese education system, including humanities, history, politics, law, education, psychology, science, technology, art and religion in MCQ format. Each question presents four candidate answers.  
\item \textbf{ARC}~\citep{DBLP:journals/corr/abs-1803-05457}: A dataset with natural, grade-school science questions (authored for human tests) in English. It is the largest public-domain set of this kind with $7,787$ questions. Each question contains four candidate answers.  
\item \textbf{MATH}: A synthetic dataset randomly generated by a script with simple mathematical questions in English. Each question has four candidate answers.
\end{enumerate}

\begin{table}
\footnotesize
\centering
\begin{tabular}{lrrl}
\toprule
 \textbf{Dataset} & \textbf{Size}  &  \textbf{Lang.} & \textbf{Format}\\
\midrule
CARE-MI  & $344$   & ZH   & MCQ/LFGQ/TFQ \\
M3KE  & $299$ &ZH   & MCQ    \\
ARC   & $291$ & EN & MCQ\\
MATH   & $300$ & EN & MCQ \\
\midrule
Total      & $1,234$ & -  & - \\
\bottomrule
\end{tabular}
\caption{Summary of the evaluation datasets. \textbf{Lang.} stands for the language of the datasets.}
\label{tab:2}
\end{table}

As shown in Table \ref{tab:2}, we ensure data balance by using a similar number of samples from each dataset.

For the first research question (\S \ref{Preference}), we use all the datasets: CARE-MI, M3KE, ARC, and MATH. In the second research question (\S \ref{cp}), we use the CARE-MI dataset for the direct output (\S \ref{do}) and token logits analysis (\S \ref{tp}) as it is the only dataset offering the three different QA formats. The ARC dataset is used on the embedding space analysis (\S \ref{e}), wherein we extend its MCQs to LFGQs by not presenting the candidate answers to the LLMs. 

The selection of benchmarks is guided by three specific criteria: (1) Source diversity: we aim to conduct our analyses across different domains. (2) Language: We presume language is a potential factor influencing LLMs evaluation performance, so we conduct experiments on two high-resource languages, i.e., Chinese and English. (3) Performance-level: By incorporating benchmarks with varying levels of LLMs demonstrated performance, we aim to better understand how model proficiency influences results in the different QA formats. Additionally, in the token logits space (\S \ref{tp}), we investigate changing the number of candidate answers to explore their impact on the expected calibration error. 

\paragraph{Prompt design} 

To encourage the generation of concise responses, we provide LLMs with prompts both prior to (\textit{pre-prompt}) and following (\textit{post-prompt}) each question in any dataset format. Additionally, for MCQs, each question is accompanied by four candidate options, with only one being correct. The pre-prompt for MCQs is ``\textit{Please select a correct option}", while the post-prompt ``\textit{Only one option can be selected. No explanation is allowed}". For LFGQs, the post-prompt is ``\textit{Just answer in one sentence}", and for TFQs, it is ``\textit{Just answer `yes' or `no'}". We find that this prompt design facilitates the generation of brief content, aiding subsequent accuracy evaluation (\S \ref{do}) and automatic confidence calculation (\S \ref{tp}).

\section{Are LLMs sensitive to the order of candidate answers?}
\label{Preference}
\begin{figure}
\begin{center}
\includegraphics[scale=0.31]{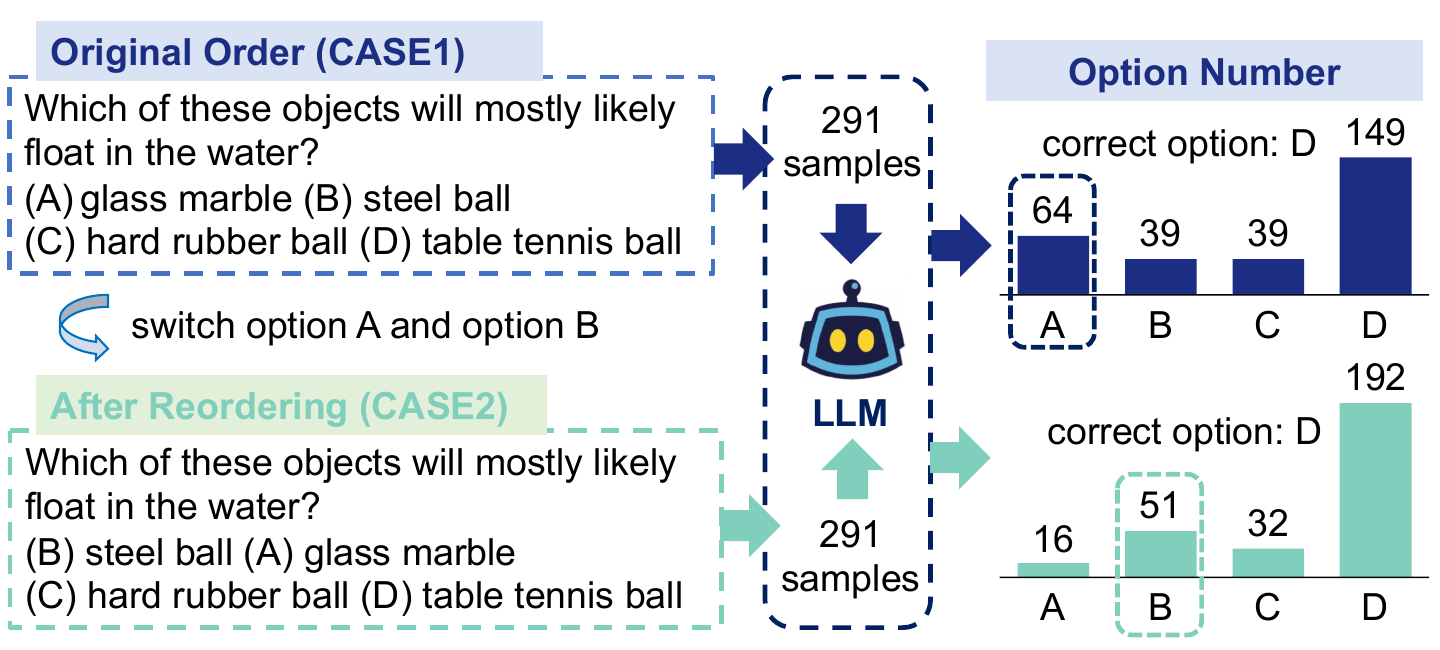} 
\caption{Example of order sensitivity experiments, in which the correct answer is D. In the ARC dataset, when predicting a wrong option (A, B, or C), the LLM prefers the option located in the first position. }
\label{fig:1}
\end{center}
\end{figure}

We first investigate how the arrangement of candidate answers in MCQs datasets affects the evaluation of LLMs.
We find that LLMs consistently exhibit a strong preference for specific positions when presented with options in different orders, as illustrated in Figure~\ref{fig:1}. 

\subsection{Order Sensitivity}
\label{st}
To check whether there are significant differences in LLMs' answers when the candidate options are arranged in a different order, we employ the chi-squared test~\citep{mchugh2013chi}. To isolate the influence of the correct answer, we designate \textit{option D} as the only correct option for all the questions. Then, we establish two scenarios: in \textbf{CASE1}, the option order is 'ABCD', and in \textbf{CASE2}, it is 'BACD'. Importantly, when arranging the option order, we also rearrange the contents and positions of each candidate option accordingly, rather than simply altering the numbering, as shown in Figure ~\ref{fig:1}.

In the chi-squared test, we set the null hypothesis, $H_0$, stating that the responses in \textbf{CASE1} and \textbf{CASE2} originate from the same distribution.
The chi-squared statistic is calculated as
\begin{equation}
    X^2 =\sum \limits_{i=0}^{N} \frac{(O_i-R_i)^2}{R_i},
\end{equation}
where $\sum \limits_{i=0}^{N}$ is the sum of $N$ candidate options , $O_i$ the  frequency of each option in \textbf{CASE1}, and $R_i$ the frequency of each option in \textbf{CASE2}. With the significance test, we can get the p-value $P$ from the chi-squared probabilities based on the chi-squared statistic and the degrees of freedom, $N-1=3$. Additionally, we can calculate the accuracy gap, which represents the difference between the original accuracy and the accuracy after reordering. 

\begin{table}
\renewcommand{\arraystretch}{1.1}
\setlength{\tabcolsep}{9pt}
\begin{tabularx}{\columnwidth}{llrrr}
\toprule 
 & & \textbf{GPT3.5} & \multirow{1}{*}{\textbf{GPT4}} &\textbf{ChatGLM}  \\
\midrule 

\rowcolor[HTML]{BAD7F2}
\multicolumn{5}{l}{\textbf{CARE-MI}}  \\
& $X^2$ & $144.192$ & $15.660$ & $27.605$  \\ 
& $P$ & **$0.000$ & *$0.001$ &  **$0.000$ \\ 
& Acc & $0.203$ & $0.637$ & $0.378$\\ 
& Gap & \textcolor{RoyalBlue}{$-0.043$} & \textcolor{RoyalBlue}{$-0.029$} & \textcolor{RoyalBlue}{$-0.116$}\\ 

\rowcolor[HTML]{BAD7F2}
\multicolumn{5}{l}{\textbf{M3KE}}  \\
& $X^2$ & $90.308$ & $20.829$ & $12.377$  \\ 
& $P$ & **$0.000$ & **$0.000$ &  **$0.006$ \\ 
&Acc & $0.381$ & $0.632$ & $0.411$\\ 
&Gap & \textcolor{RoyalBlue}{$-0.030$} & \textcolor{RoyalBlue}{$-0.017$}& \textcolor{Melon}{$+0.014$}\\ 

\rowcolor[HTML]{BAD7F2}
\multicolumn{5}{l}{\textbf{ARC}}  \\
& $X^2$ & $36.515$ & $2.681$ & $10.511$  \\ 
& $P$ & **$0.000$ & \textbf{$0.443$} & *$0.015$ \\ 
& Acc& $0.512$ & $0.935$ & $0.553$\\ 
& Gap& \textcolor{Melon}{+$0.148$} & \textcolor{RoyalBlue}{-$0.031$} & \textcolor{RoyalBlue}{-$0.116$}\\ 

\rowcolor[HTML]{BAD7F2}
\multicolumn{5}{l}{\textbf{MATH}}  \\
& $X^2$ & $25.129$ & $4.513$ & $90.566$  \\ 
& $P$ & **$0.000$ & \textbf{$0.211$} &  **$0.000$ \\ 
& Acc & $0.597$ & $0.780$ & $0.480$\\ 
& Gap & \textcolor{Melon}{+$0.000$} & \textcolor{RoyalBlue}{-$0.023$} & \textcolor{RoyalBlue}{-$0.043$}\\ 
\bottomrule   
\end{tabularx}
\caption{LLMs' order sensitivity results. The rearrangement of options makes LLMs output different answers.* indicates $P<0.05$, ** indicates $P<0.001$, and bold indicates larger than significance level $\alpha$. 
}
\label{tab:3}
\end{table}

Results are shown in Table~\ref{tab:3}, from which we note the following observations:

\begin{enumerate} 

\item There is a considerable disparity in LLMs' outputs across the two scenarios. Except for two instances,\footnote{GPT4 model on the ARC ($X^{2}=2.681$, $P= 0.443$) and the MATH ($X^{2}= 4.513$, $P= 0.211$) datasets.} all the results have a p-value $P < 0.05$, rejecting the null hypothesis and implying that the distribution of answers predicted by the model varies significantly when options A and B are interchanged. This indicates that the order of options significantly influences LLMs' predictions in MCQs datasets.

\item Among the GPT family, the rearrangement of options has a more pronounced effect on GPT-3.5-turbo, with bigger accuracy gaps, than on GPT-4. 

\item Higher accuracy can mitigate significant differences in the order arrangement to some extent. Results from GPT-4 on the ARC and the MATH datasets indicate that high accuracies ($\geq 0.780$) can lead to not rejecting the null hypothesis. 

\item There is no evident correlation between the accuracy gap and the original accuracy. A higher accuracy does not necessarily imply a lower gap between the two scenarios.
\end{enumerate}

\subsection{Pattern Decomposition}
\label{pe}
Next, we further explore the pattern decomposition of LLMs to investigate potential patterns underlying their sensitivity to order. We propose the following two hypotheses for exploration: (1) LLMs may have different positional preferences due to their different model bases; (2) LLMs may have different positional preferences depending on whether they have previously memorized the contents of the datasets.

\begin{table*}
\centering
\begin{tabularx}{0.85\linewidth}{lccrrrcrrrcrrr}
\toprule 
& & & \multicolumn{3}{c}{\textbf{GPT-3.5-turbo}} & & \multicolumn{3}{c}{\textbf{GPT-4}} & & \multicolumn{3}{c}{\textbf{ChatGLM-6B}}\\
\cline{4-6}
\cline{8-10}
\cline{12-14}
\textbf{Dataset}  &\textbf{CASE} & & \textbf{1st} & \textbf{2nd} & \textbf{3rd} & & \textbf{1st} & \textbf{2nd} & \textbf{3rd} & & \textbf{1st} & \textbf{2nd} & \textbf{3rd}\\ 
\midrule 

\multirow{2}{*}{CARE-MI}& C1 & & \cellcolor{Melon!41} $41$ & \cellcolor{Melon!33} $33$ & \cellcolor{Melon!26} $26$ & & \cellcolor{Melon!44} $44$ & \cellcolor{Melon!37} $37$ & \cellcolor{Melon!19} $19$ & & \cellcolor{Melon!37} $37$ &\cellcolor{Melon!42} $42$ & \cellcolor{Melon!21} $21$\\ 

& C2  & & \cellcolor{Melon!83} $83$ & \cellcolor{Melon!8} $8$& \cellcolor{Melon!9} $9$ & & \cellcolor{Melon!61} $61$ & \cellcolor{Melon!26} $26$ & \cellcolor{Melon!13} $13$ & & \cellcolor{Melon!61} $61$ & \cellcolor{Melon!23} $23$ & \cellcolor{Melon!16} $16$\\ 

\multirow{2}{*}{M3KE} & C1 & &\cellcolor{Melon!25} $25$ &\cellcolor{Melon!35} $35$ & \cellcolor{Melon!40} $40$ & & \cellcolor{Melon!49} $49$ &\cellcolor{Melon!26} $26$ &\cellcolor{Melon!25} $25$ & & \cellcolor{Melon!33} $33$ &\cellcolor{Melon!39} $39$ &\cellcolor{Melon!28} $28$\\

& C2 & & \cellcolor{Melon!75} $75$  &\cellcolor{Melon!1} $1$ & \cellcolor{Melon!24} $24$ & & \cellcolor{Melon!49} $49$  &\cellcolor{Melon!31} $31$ &\cellcolor{Melon!30} $30$ & & \cellcolor{Melon!46} $46$  &\cellcolor{Melon!32} $32$ &\cellcolor{Melon!22} $22$\\

\multirow{2}{*}{ARC} & C1 & & \cellcolor{Melon!47} $47$ & \cellcolor{Melon!27} $27$ &\cellcolor{Melon!26} $26$ & & \cellcolor{Melon!90} $37$ &\cellcolor{Melon!10} $31$ &\cellcolor{Melon!30} $32$ & &\cellcolor{Melon!33} $33$ &\cellcolor{Melon!28} $28$ &\cellcolor{Melon!39} $39$\\

& C2 & &\cellcolor{Melon!52} $52$  &\cellcolor{Melon!16} $16$ &\cellcolor{Melon!32} $32$ & & \cellcolor{Melon!21} $21$  &\cellcolor{Melon!36} $36$ &\cellcolor{Melon!43} $43$ & & \cellcolor{Melon!33} $33$  &\cellcolor{Melon!36} $36$ &\cellcolor{Melon!31} $31$\\

\multirow{2}{*}{MATH} & C1 & &\cellcolor{Melon!22} $22$ &\cellcolor{Melon!60} $60$ &\cellcolor{Melon!18} $18$ & & \cellcolor{Melon!41} $41$ &\cellcolor{Melon!30} $30$ &\cellcolor{Melon!29} $29$ & & \cellcolor{Melon!17} $17$ &\cellcolor{Melon!69} $69$ &\cellcolor{Melon!14} $14$\\

& C2 & &\cellcolor{Melon!61} $61$  &\cellcolor{Melon!3} $3$ &\cellcolor{Melon!36} $36$ & & \cellcolor{Melon!47} $47$  &\cellcolor{Melon!28} $28$ &\cellcolor{Melon!25} $25$ & & \cellcolor{Melon!22} $22$  &\cellcolor{Melon!57} $57$ &\cellcolor{Melon!11} $11$\\
\bottomrule    
\end{tabularx}
\caption{Pattern decomposition results. C1 refers to \textbf{CASE1}, and C2 to \textbf{CASE2}. The numbers indicate the percentage (\%) of incorrect options (A, B, or C) that each model selects for each position (1st position, 2nd position, and 3rd position). Deeper \colorbox{Melon}{background} indicates a higher preference for that position.}
\label{tab:5}
\end{table*}
We use the same LLMs as in \S\ref{st}, as they can provide concise answers to the questions and come from different model bases. Regarding the datasets, CARE-MI, M3KE, and ARC are derived from website documents, while the MATH dataset, synthetically generated by us, ensures that the LLMs have not been exposed to identical questions during training. Results are presented in Table~\ref{tab:5}, from which we can extract the following conclusions:

\begin{enumerate} 
\item Within the GPT family, GPT-3.5-turbo and GPT4 exhibit different behavior. 
When predicting incorrect options (A, B, or C), GPT4 shows a stronger inclination towards the option positioned first compared to GPT-3.5-turbo. Specifically, when option B is presented first, GPT-3.5-turbo tends to lean towards selecting option B. Furthermore, ChatGLM-6B showcases a certain preference for the first two options.

\item The behavior of the LLMs remains consistent across datasets originating from different languages and sources. Hence, we can conclude that the dataset's language or source, regardless of whether the models were previously exposed to them or not, is not the underlying cause of the models' positional preferences.
\end{enumerate} 

\subsection{Yes, LLMs are sensitive to ordering}
Our experiments showed that the order of candidate answers in MCQs significantly impacts LLMs outputs. GPT-3.5-turbo and GPT4 exhibited different preferences, 
while ChatGLM-6B showed a
certain preference for the first two positions. In addition, the positional preferences in each LLM seemed to remain
consistent across datasets originating from different languages and sources. 

These findings are problematic because they reveal potential biases and inconsistencies in LLM outputs, which can affect the reliability and accuracy of their responses. Failure to understand and address these preferences may lead to biased recommendations, inaccurate information retrieval, and flawed decision-making. It is important to develop methods to mitigate these effects, as well as to investigate evaluation protocols that are less impacted by positional preferences. In light of these observations, in the next section, we compare MCQs and LFGQs evaluation methods.

\section{Multiple Choice vs\\Long Form Generation}
\label{cp}
To compare QA evaluation formats and gain a deeper understanding of LLMs evaluation protocols, we expose several LLMs to the same questions presented in different formats. Then, we analyze and compare the results in three spaces:
the direct output space (\S \ref{do}), the token logits space (\S \ref{tp}), and the embedding space (\S \ref{e}).

\subsection{Direct Output}
\label{do}
In the direct output space, which refers to the responses generated by the LLMs, accuracy is one of the most common evaluation metrics used for benchmarking purposes and performance assessment. The difference in accuracy between the MCQs and LFGQs formats is the first aspect we consider (\S \ref{at}). Additionally, we study the relationship between consistency and accuracy (\S \ref{ca}) by exploring whether LLMs, if familiar with a particular concept, tend to generate responses that are similar and encompass consistent factual information~\citep{DBLP:conf/emnlp/ManakulLG23}.

\subsubsection{Accuracy}
\label{at}
\begin{figure}
\begin{center}
\includegraphics[scale=0.34]{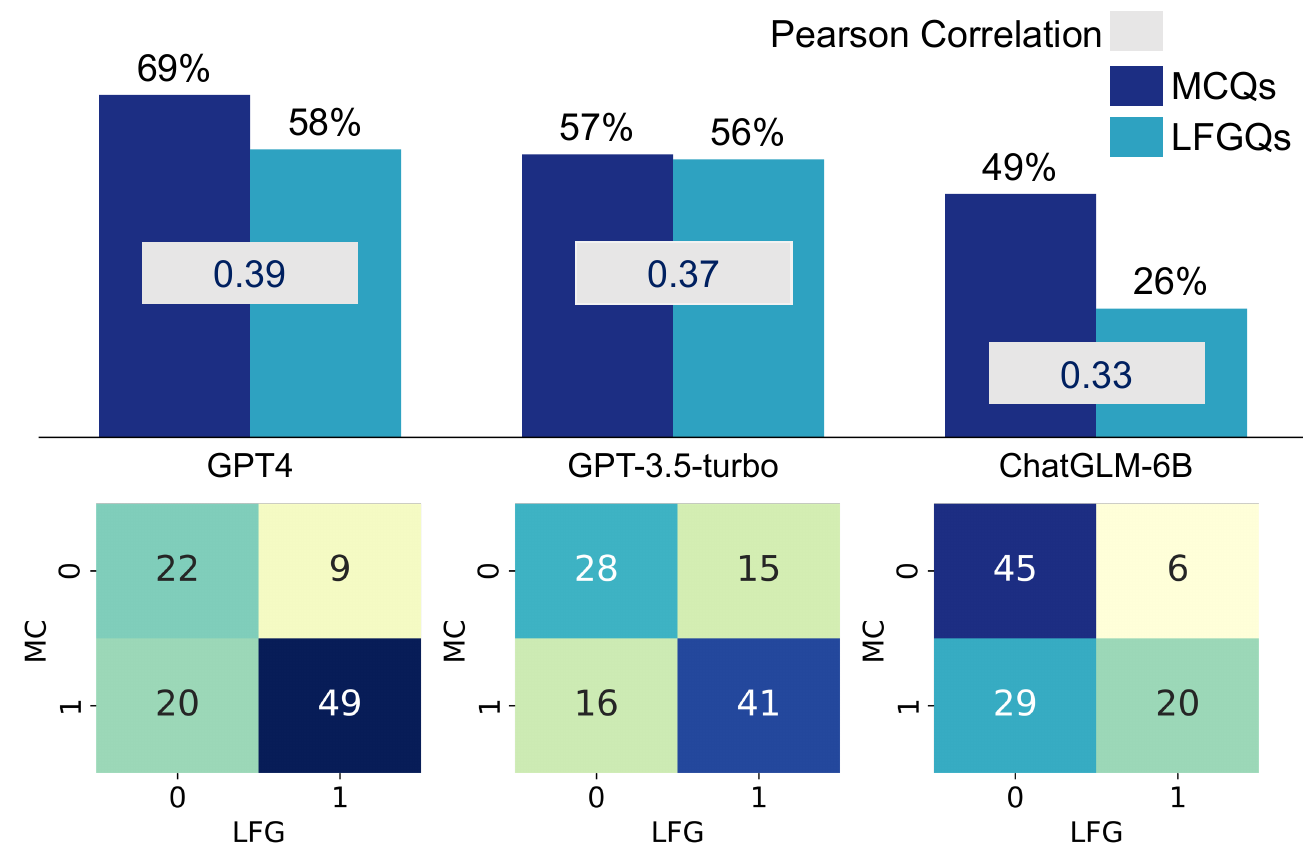} 
\caption{Comparison between MCQs and LFGQs on the CARE-MI dataset. Top: accuracy and pearson correlation. Bottom: MCQ vs LFGQs matrix.}
\label{fig:2}
\end{center}
\end{figure}

We randomly select $100$ samples from the CARE-MI dataset and evaluate GPT4, GPT-3.5-turbo, and ChatGLM-6B on them. For MCQs, accuracy is computed as usual, i.e. if the predicted answer matches the ground truth, it is considered correct. For LFGQs,  accuracy is determined through human evaluation, with $0$ denoting an incorrect response and $1$ a correct one. Figure \ref{fig:2} (top) compares the accuracy between MCQs and LFGQs across the three LLMs. Notably, the accuracies of MCQs are consistently higher than those of LFGQs. This difference can be attributed to the fact that MCQs offer candidate options, facilitating the prediction task. To delve deeper into the analysis, in Figure \ref{fig:2} (bottom), we visualize a matrix in which, for the same question, there are four scenarios: 1) the response is correct in both formats, 2) the response is incorrect in both formats, 3) the response is correct in MCQs but incorrect in LFGQs, and 4) the response is correct in LFGQs but incorrect in MCQs. Results show that there are a relatively large number of questions where the LLMs can respond correctly in MCQs, but fail in the LFGQs format. Furthermore, we quantify the differences in accuracy produced by the two formats with Pearson correlation coefficients. The obtained values are remarkably low: $0.39$ for GPT4, $0.7$ for GPT-3.5-turbo, and $0.33$ ChatGLM-6B, clearly indicating that different versions of the same question yield different answers from the LLM.

\subsubsection{Consistency}
\label{ca}
Next, we explore the relationship between consistency and accuracy. Consistency stands for the degree to which the LLMs provide the same answer when asked the same question multiple times. For example, an answer of `AAAAB' is considered more consistent than `BCDAB' when presented with the same question five times. To conduct this evaluation, we first define the quantitative measures for consistency and accuracy in a sequence of responses to a repeated question.

Formally, let $\mathcal{A} = \{A_1, A_2, ..., A_D\}$ be a sequence of answers, where a model is queried $D$ times. Each answer $A_i \in \mathcal{A}$ is selected from a set of $N$ unique options $\mathcal{O} = \{Opt_1, Opt_2, ..., Opt_N\}$. From $\mathcal{A}$, we derive a count sequence

$\mathcal{C} = \{\textsc{count}(Opt_1), \textsc{count}(Opt_2), ..., \textsc{count}(Opt_N)\}$, where $\textsc{count}(Opt_i)$ represents the number of occurrences of $Opt_i$ in $\mathcal{A}$, marking $Opt_{\max}$ as the option with the largest $\textsc{count}(Opt_i)$. We define sequence consistency $K$ as
\begin{equation}
\begin{split}
K(\mathcal{A}) & = \frac{1}{D}   \textsc{count}(Opt_{max}) \\
&+ \frac{1}{D} \sum \limits_{Opt_i \neq  Opt_{\max}} \max(0, \textsc{count}(Opt_i)-1).
\end{split}
\end{equation}

As for accuracy, if $A_{ref} \in \mathcal{O}$ is the correct answer,
the accuracy for sequence $\mathcal{A}$ can be defined as
\begin{equation}
\text{Acc}(\mathcal{A}) = \frac{1}{D}  \textsc{count}(A_{ref}).
\end{equation}

In the experiments, we use the same samples as in \S \ref{at} and repeat each question five times for each LLM. To compute consistency and accuracy on LFQs, we manually group the long-text generated answers into options so that answers with similar meanings are grouped together under the same option. We also explore the impact of different temperatures, which is the parameter that controls the degree of randomness of the generated text, by using values $0$, $0.5$, and $1$. 

\begin{figure}
\begin{center}
\includegraphics[scale=0.6]{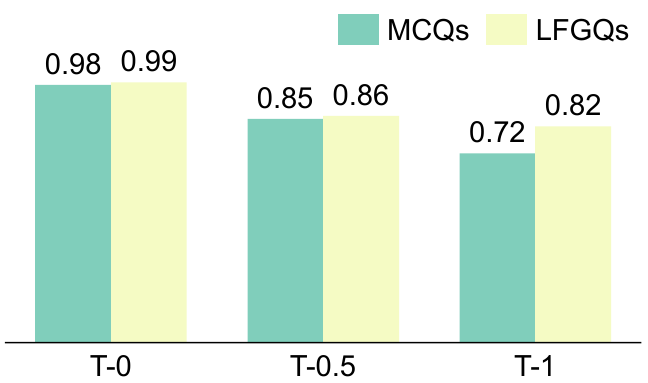} 
\caption{GPT-3.5-turbo's consistency on MCQs and LFGQs with different temperatures.}
\label{fig:3}
\end{center}
\end{figure}

Figure \ref{fig:3} shows GPT-3.5-turbo's consistency for MCQs and LFGQs across the three temperature values. Both formats tend to be consistent in their answers. Even when the temperature is increased, consistency does not decrease notably. Between the two formats, LFGQs tend to be more consistent than MCQs. We also calculate the Pearson correlation coefficient between consistency and accuracy. In the case of MCQs, the Pearson correlation coefficient is $0.32$, while for LFGQs, the coefficient reaches $0.416$, implying that higher consistency does not necessarily mean more correct. Our findings suggest that a higher level of consistency indicates a sharper probability distribution of specific knowledge, but it does not guarantee the correctness of the knowledge. Unlike SelfCheckGPT~\citep{DBLP:conf/emnlp/ManakulLG23}, which leverages the idea that the higher the consistency, the higher the correctness, we do not find a direct relationship between consistency and accuracy. 
We believe this is due to the knowledge required to answer the evaluation dataset. While SelfCheckGPT is evaluated on information from famous individuals~\cite{DBLP:conf/emnlp/LebretGA16}, we use specialized professional medical datasets.

\subsection{Token Logits}
\label{tp}
To compare MCQs and LFGQs in the token logits space, which is the space of predicted probabilities, we rely on two techniques: unified confidence calculation and expected calibration error.

\subsubsection{Unified confidence calculation}
One of the mainstream approaches used to analyze why LLMs select specific options when answering MCQs is through token logits~\citep{DBLP:conf/emnlp/ManakulLG23}. GPT-3.5-turbo, for instance, can generate log probabilities for the most probable tokens associated with each output token.\footnote{\url{https://platform.openai.com/docs/guides/gpt}}
However, while there are formulas to calculate confidence for multiple options~\citep{DBLP:journals/tacl/JiangADN21,holtzman2021surface,DBLP:journals/tmlr/LinHE22}, direct utilization of token probability calculations for comparing MCQs with LFGQs is not straightforward.

Since our goal is to compare MCQs and LFGQs, we follow~\cite{DBLP:journals/tacl/JiangADN21} and propose a unified confidence calculation applicable to the three QA formats: MCQs, LFGQs, and TFQs.
Let us assume an input question $q$ that makes a LLM generate the set of answers $\mathcal{A}$.

Each answer $A_i \in \mathcal{A}$ contains $|A_i|$ tokens. Each token, denoted as $t_k$, with $1 \leq k \leq |A_i|$, has a corresponding autoregressive token log probability $P_{log}(t_k|q,t_{<k})$. We first compute the average token log probability of each answer $A_i$ as
\begin{equation}
    P_{avg}(A_i|q) = \frac{
    {\sum_{i=1}^{|A_i|}
    }P_{log}(t_k|q,t_{<k})}
    {\max(1,|A_i|)}
    .
\end{equation}

From the initial set of answers $\mathcal{A}$, which may contain duplicates, we consolidate them into $z$ unique answers, denoted as $\mathcal{A}^{uni} = \{A_{1}^{uni}, A_{2}^{uni}, ..., A_{z}^{uni}\}$, where $z \leq D$. For each unique answer, we select the highest log probability observed for any instance of that answer in $\mathcal{A}$, denoted as $P_{log}^{highest}(A_{i}^{uni}|q)$. Subsequently, we rank the first $W$ unique answers, where $W \leq z$,\footnote{$W=4$ in our experiments.} from $\mathcal{A}^{uni}$ in descending order by their frequency and the corresponding $P_{log}^{highest}(A_{i}^{uni}|q)$, to filter out excessively similar responses and maintain the diversity in the unique answers. We calculate the standardized confidence for the first $W$ answers as
\begin{equation}
   C_N(A_{w}^{uni}) = \frac{e^{p^{higest}_{avg}}(A_{w}^{uni}|q)}{\sum^{W}_{w=1}e^{p^{higest}_{avg}}(A_{w}^{uni}|q)}.
\end{equation}

Finally, For MCQs, we use regularization matching to combine first $W$ answers with four candidate options, and get the final label ($0$ or $1$) with the sum of corresponding standardized confidence. For LFGQs, we directly get the final label ($0$ or $1$) according to standardized confidence for the first $W$ answers and human labeling for each answer.

\subsubsection{Expected Calibration Error}

After obtaining the unified confidence, we compute model calibration~\citep{gupta2006model,ahmed2020models} to test whether
a LLM exhibits good calibration across different dataset evaluation formats. A well-calibrated model should provide confidence (i.e., logit) estimates that closely match the actual probability of the correctness of the answer. Inaccurate predictions should correspond to low confidence (i.e, logit) values, whereas accurate predictions should yield high confidence (i.e., logit) values.

In practice, we employ a commonly used metric known as expected calibration error (ECE)~\citep{DBLP:conf/icml/Niculescu-MizilC05} to assess the alignment of confidence and accuracy. ECE is computed as the weighted average of the difference between the accuracy and confidence. 
To measure confidence quantitatively, we divide the $[0, 1]$ interval into multiple bins. Each sample falls into one of these bins based on the model's predicted results. The average model confidence is calculated in each bin, and then compared with the average accuracy of the sample real label in the bin. The absolute value of these two differences can measure the model's confidence. A larger difference indicates lower model confidence. Formally,

\begin{equation}
    \text{ECE} = \sum_{b=1}^{\mathcal B} \frac{|n_b|}{N}|\text{acc}(b)-\text{conf}(b)|.
\end{equation} %
where $b$ represents the $b$-th bin, $\mathcal B$ represents the total number of bins, $n_b$ represents the number of samples in the $b$-th bin, $\text{acc}(b)$ represents the average value of the true label of the sample in the $b$-th bin, $\text{conf}(b)$ represents the average value of the model prediction probability in the $b$-th bin. In our experiments, we set $\mathcal B = 100$.

\subsubsection{Results}
\begin{figure*}
\begin{center}
\includegraphics[scale=0.53]{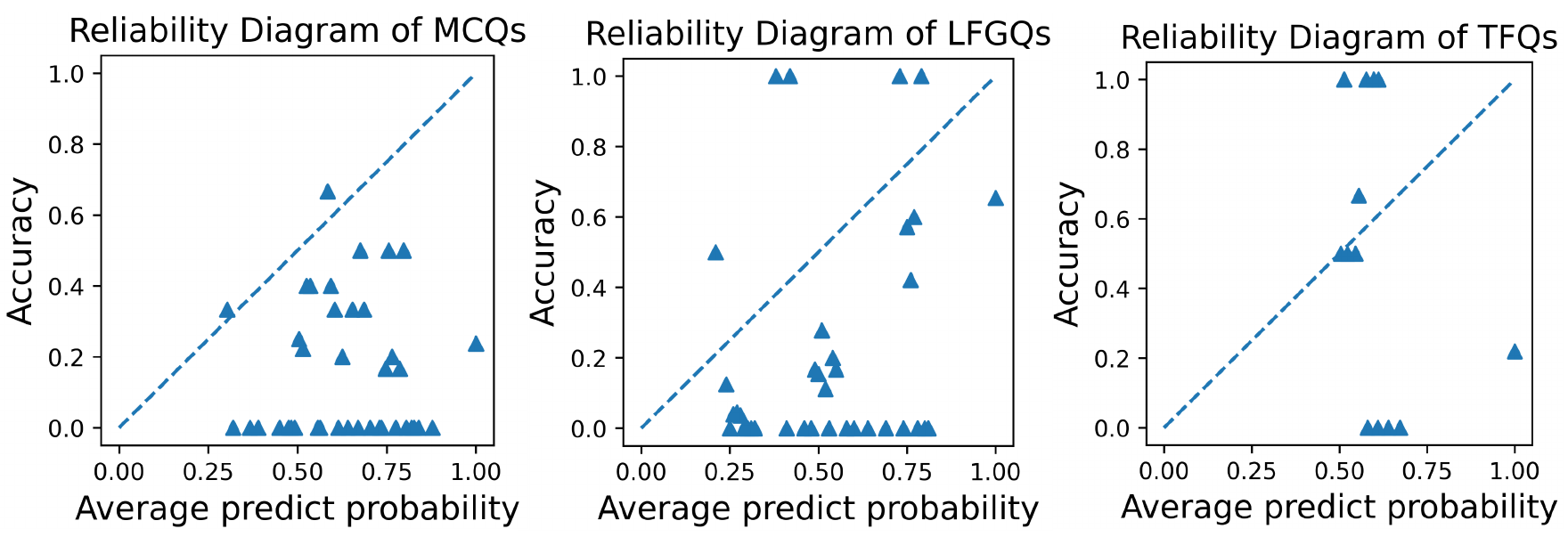} 
\caption{Reliability diagrams for MCQs, LFGQs, and TFQs on the CARE-MI dataset.}
\label{fig:4}
\end{center}
\end{figure*}

Within the CARE-MI dataset, we use the three QA formats, MCQs, LFGQs, and TFQs, to compute ECE and reliability. We use confidence scores and true labels to draw reliability diagrams as in~\citep{DBLP:journals/corr/abs-2203-08958}. A reliability diagram closely aligning with the identity line suggests good model calibration, while a significant deviation indicates poor calibration. Results are shown in Figure \ref{fig:4} and in Table \ref{tab:7}. LLMs operating on MCQs exhibit the poorest calibration and highest ECE compared to the other two formats. This suggests that the LLMs' predictions in MCQs are not accurately aligned with the true probability of correct answers,  indicating overconfidence in their responses. Additionally, we observe that the ECE in TFQs ($0.276$), which contain only two candidate options, is lower than in MCQs ($0.426$), which have four candidate options. To investigate the impact of the number of candidate answers, we conduct experiments by varying the number of options in MCQs across the four datasets. We analyze whether the number of options and the domain of each dataset affect ECE. Throughout these experiments, we maintain the correct answer consistently positioned as the last option. As depicted in Table \ref{tab:7}, we do not find a clear correlation between these factors and ECE, meaning that the number of options and the domain do not seem to influence LLM's performance.

\begin{table}
\setlength{\tabcolsep}{4pt}
\centering
\begin{tabularx}{\columnwidth}{lrrrrr}
\toprule

\textbf{Format} & \textbf{CA} & \textbf{CARE-MI} & \textbf{M3KE} & \textbf{ARC} & \textbf{MATH} \\

\hline
MCQ & $4$ & $0.426$ & $0.317$ & $0.492$ & $0.281$ \\
MCQ & $3$ & $0.329$ & $0.364$ & $0.382$ & $0.259$  \\
MCQ & $2$ & $0.414$ & $0.427$ & $0.280$ & $0.257$\\
LFGQ & - & $0.304$ & - & - & - \\

TFQ & $2$ & $0.276$ & - & - & - \\
\bottomrule
\end{tabularx}
\caption{GPT-3.5-turbo's ECE for different formats and number of candidate answers. CA stands for the number of candidate answers.}
\label{tab:7}
\end{table}

\subsection{Embeddings}
\label{e}
Up to this point, our analysis reveals that the misalignment between MCQs and LFGQs answers is evident in both the direct output (\S \ref{do}) and the token logits (\S \ref{tp}). Next, we investigate whether this difference is also manifested in the embedding space derived from the hidden states of the models~\citep{DBLP:conf/iclr/BurnsYKS23}. We also explore how the embeddings behave under different question formats and models.
The technique proposed by \citet{DBLP:journals/corr/abs-2306-03341} enables the extraction of hidden outputs from the model by collecting the heads of the attention blocks, and use these heads as index to obtain the hidden outputs of each layer from the model.\footnote{\url{https://github.com/davidbau/baukit}} We utilize the hidden outputs of the last token in the input. To thoroughly investigate the distinctions in the embedding space between MCQs and LFGQs themselves as much as possible, unlike the prompt design in the previous experiments, we only set a post-prompt for MCQs, and LFGQs do not contain any prompts.
Finally, the hidden outputs have information on the number of input samples, the number of hidden layers, the number of attentions, and the dimensions of heads. Refer to Table \ref{tab:11} in the Appendix (\S \ref{App}) for more details.
We randomly select $40$ samples from the ARC dataset, each with MCQs and LFGQs formats, and plot t-SNE~\citep{van2008visualizing} representations of the hidden embeddings in each layer. Figure \ref{fig:6} shows the visualizations for Llama-2-7b-chat-hf. Other model visualizations are provided in the Appendix (\S \ref{App}). The results show that the embeddings from MCQs and LFGQs display clear separations in some layers of the hidden states. We observe a consistent trend across the various LLMs: in the initial layers, embeddings of the two formats show clear separations. However, as we progress towards the final layers, the embeddings corresponding to MCQs and LFGQs tend to become closer. Additionally, in certain models, the embeddings are distinctly separated in specific middle layers. For instance, in the open-llama-7b model, the embeddings exhibit clear differentiation in the 14th layer. Finally, the representation of embeddings from the same model but different sizes can vary, as shown in the embeddings of Dolly-v2-3b and Dolly-v2-7b in Figures \ref{fig:69} and \ref{fig:70} in the Appendix. 

\begin{figure}
\begin{center}
\includegraphics[scale=0.23]{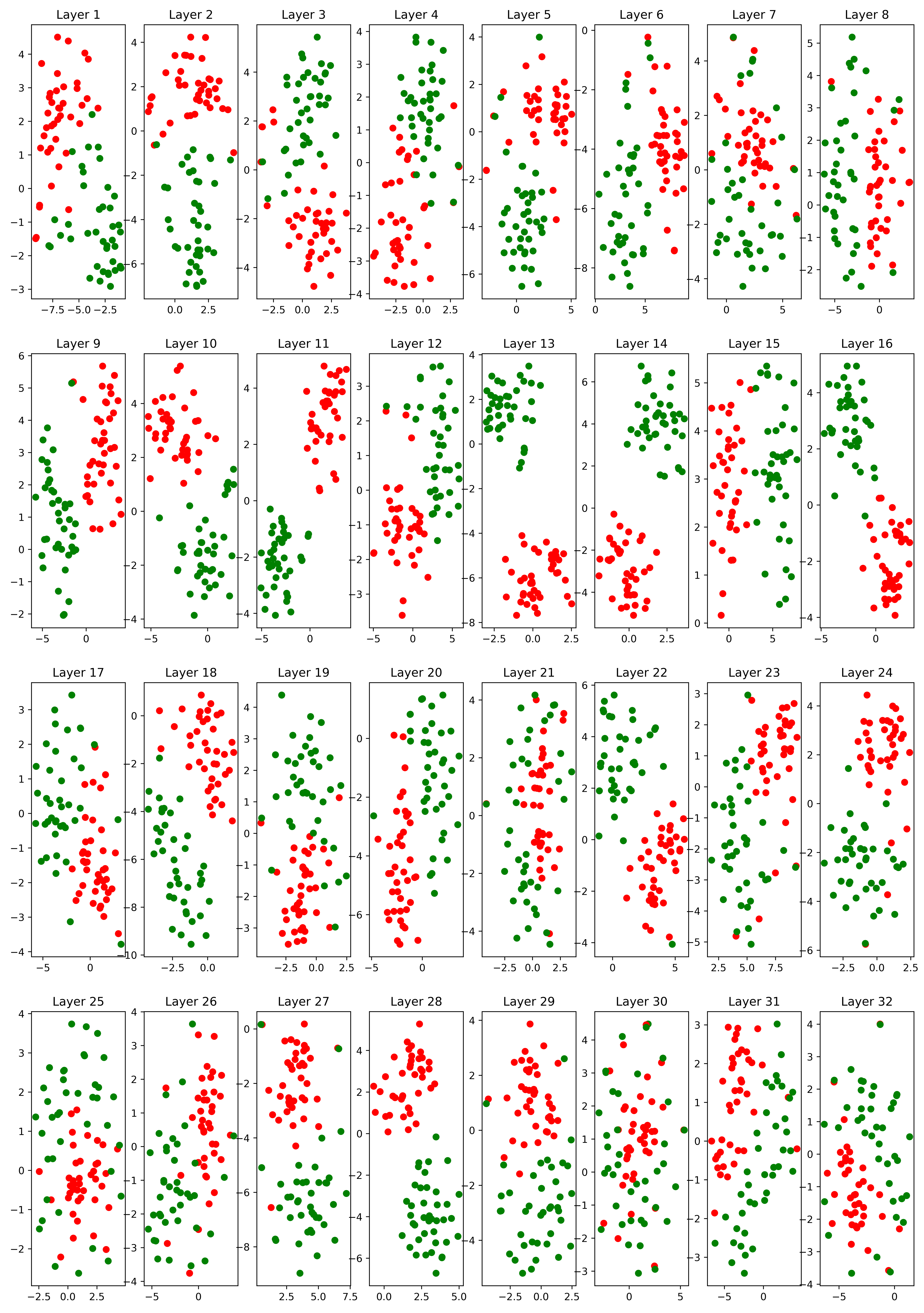} 
\caption{t-SNE visualization for each layer in Llama-2-7b-chat-hf. MCQs in red, LFGQs in green.}
\label{fig:6}
\end{center}
\end{figure}

\subsection{Different QA formats produce different answers}
Our experiments showed that different formats of a single question may result in different performances. Moreover, our results challenged the notion that greater consistency leads to higher accuracy by closely examining the relationship between them. When comparing MCQs and LFGQs in expected calibration error, prompts from MCQs were the most overconfident in their predictions. Finally, in the embedding space, MCQs and LFGQs representations were clearly separated in some layers of the hidden states.

\section{Related Work}

\subsection{QA Benchmarks} 
QA is a prevalent evaluation method in natural language processing tasks. With the surge of LLMs, several QA evaluation benchmarks have emerged to asses models' reasoning and fact-retrieval skills~\citep{DBLP:journals/corr/abs-2211-09110,DBLP:journals/corr/abs-2307-03109,DBLP:journals/corr/abs-2306-09212,DBLP:journals/corr/abs-2306-04757}. These QA benchmarks encompass divers dataset formats, including multiple-choice questions (MCQs)~\citep{DBLP:journals/corr/abs-2102-03315,DBLP:conf/icpr/RamamurthyA22,DBLP:journals/corr/abs-2305-10263,DBLP:journals/corr/abs-2305-08322}, long-form generation questions (LFGQs)~\citep{DBLP:journals/access/ZhangZWGL18,DBLP:conf/acl/LinHE22,CARE-MI} 
and True/False questions (TFQs)~\citep{DBLP:journals/corr/abs-2305-09617}. Many existing QA evaluation benchmarks use relatively simple MCQs formats, in which models can strongly rely to formulate their answers. In addition, previous work has primarily focused on evaluations of MCQs, not considering comparisons between the different formats~\citep{DBLP:journals/tacl/JiangADN21,DBLP:journals/tmlr/LinHE22,DBLP:conf/iclr/RobinsonW23}. In this paper, we focused on conducting a comprehensive comparative analysis between MCQs and LFGQs, thereby enhancing the understanding of the drawbacks and limitations of the different evaluation methods.

\subsection{LLMs and Multiple-Choice Questions}
Previous work has underscored the sensitivity of LLMs to prompting strategies~\citep{DBLP:conf/icml/ZhaoWFK021,DBLP:journals/corr/abs-2305-09617} and positional bias~\citep{DBLP:journals/corr/abs-2305-17926}, which pose challenges to model assessment. For instance,
\cite{DBLP:journals/corr/abs-2306-05685} showed that GPT-4 tends to favor the candidate answer presented in the first position, leading to unfair evaluation results.  Additionally, \cite{DBLP:journals/corr/abs-2308-11483} observed that GPT-4 and InstructGPT\cite{DBLP:conf/nips/Ouyang0JAWMZASR22} perform differently when answer options are rearranged on various benchmarks. We expand upon prior work, which focused on a limited number of models and scenarios, to study and identify general patterns and analyze their underlying causes across diverse datasets and models.

\section{Discussion and Conclusion}
This paper focused on testing the effectiveness of MCQs evaluating LLMs. Motivated by the observation of consistent preference biases across different datasets with several LLMs, we first conducted a significance test to determine the position of the candidate answers affect LLMs' predictions, resulting in accuracy instability. More specifically, we analyzed how different LLMs have different positional preference patterns on the same dataset, while the preference positional patterns of a particular LLM remained constant across datasets from different sources. In addition,
we conducted comparative experiments between MCQs and LFGs in three different spaces to ascertain the advantages and disadvantages of each as evaluation benchmarks. 

\paragraph{Recommendations}
Based on our experiments, we offer a few suggestions for utilizing MCQs and LFGQs formats in LLM evaluation benchmarks:
\begin{enumerate}
\item The choice of QA format should be aligned with the type of knowledge being evaluated. Whereas it may be fine to use MCQs for testing general knowledge, in some professional domains—particularly those carrying legal responsibilities, such as the medical field, it is advisable to use LFGQs under human supervision to ensure a more rigorous evaluation.
\item When using MCQs for evaluating LLM, adjusting the number of options, whether decreasing or increasing them, does not necessarily enhance accuracy and confidence. However, regarding order sensitivity, reordering candidate answers for each question and repeating questions can enhance the robustness of the assessment process.
\item Our findings do not indicate a strong correlation between consistency and accuracy in LLMs responses. Therefore, we do not recommend relying on consistency as a tool to enhance performance in LLMs.
\item Given the discrepancy we found between MCQs and LFGQs results, we believe that LFGQs is the best format for evaluating LLM, as it aligns well with real-world use cases. We recommend prioritizing LFGQs format and evaluating LLM from various perspectives, including correctness, completeness, relevance, and interpretability.
\end{enumerate}

We hope that the results presented in the paper and the investigation about order sensitivity and comparative analyses between MCQs and LFQs can inspire future research to improve evaluation benchmarks for LLMs.

\section*{Acknowledgment}
This work was partly supported by JSPS KAKENHI No.JP22K12091 and the State Key Program of National Nature Science Foundation of China No.61936001.
\newpage
\section{Reference}\label{sec:reference}
\bibliographystyle{lrec-coling2024-natbib}
\bibliography{lrec-coling2024-example}

\newpage
\section{Appendix}
\label{App}
\begin{figure}[H]
\begin{center}
\includegraphics[scale=0.03]{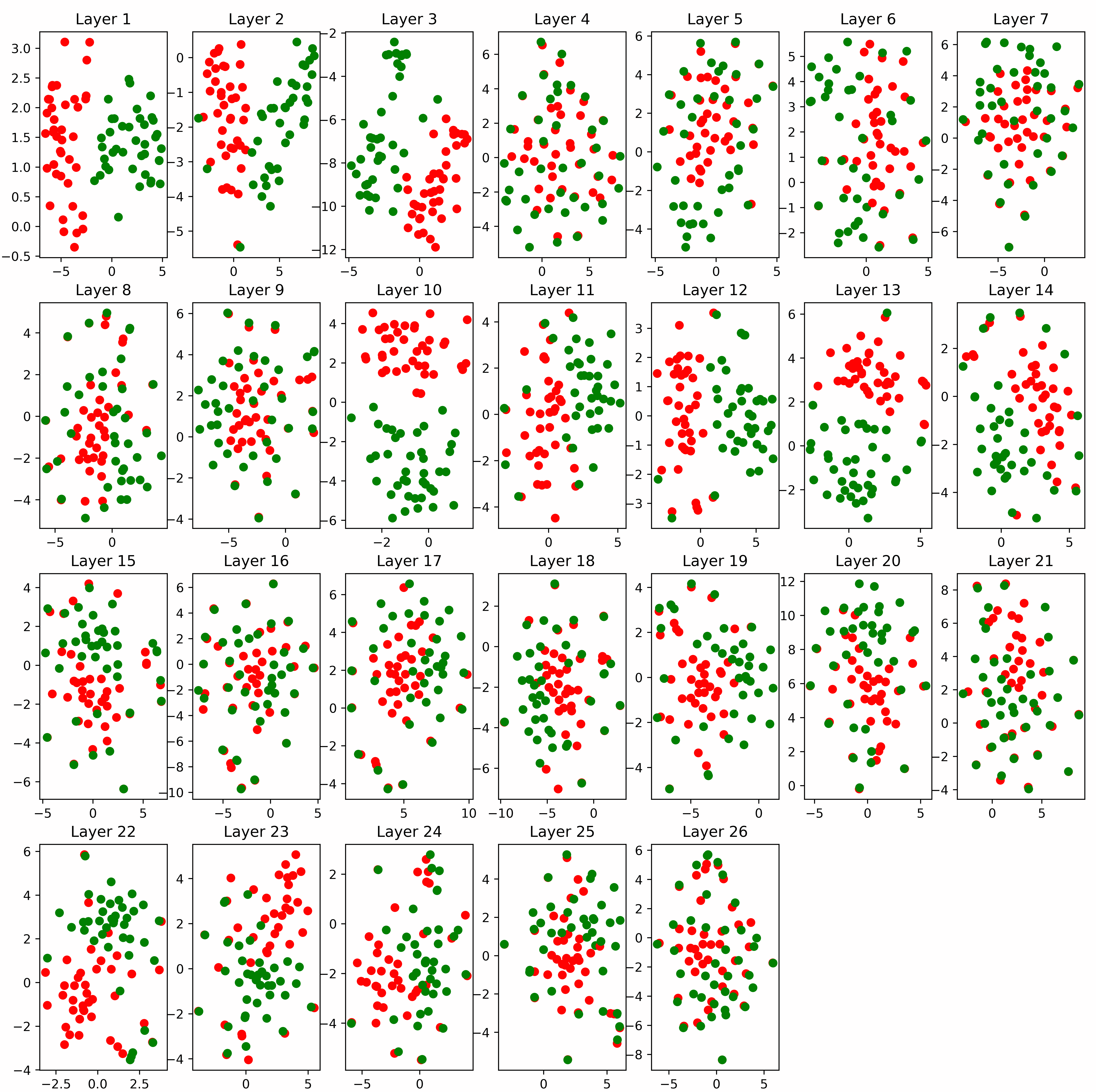} 
\caption{The visualization of t-SNE for each layer in the model Open-llama-3b. The red samples are MCQs, and the samples in green are LFGQs.}
\label{fig:66}
\end{center}
\end{figure}

\begin{figure}[H]
\begin{center}
\includegraphics[scale=0.03]{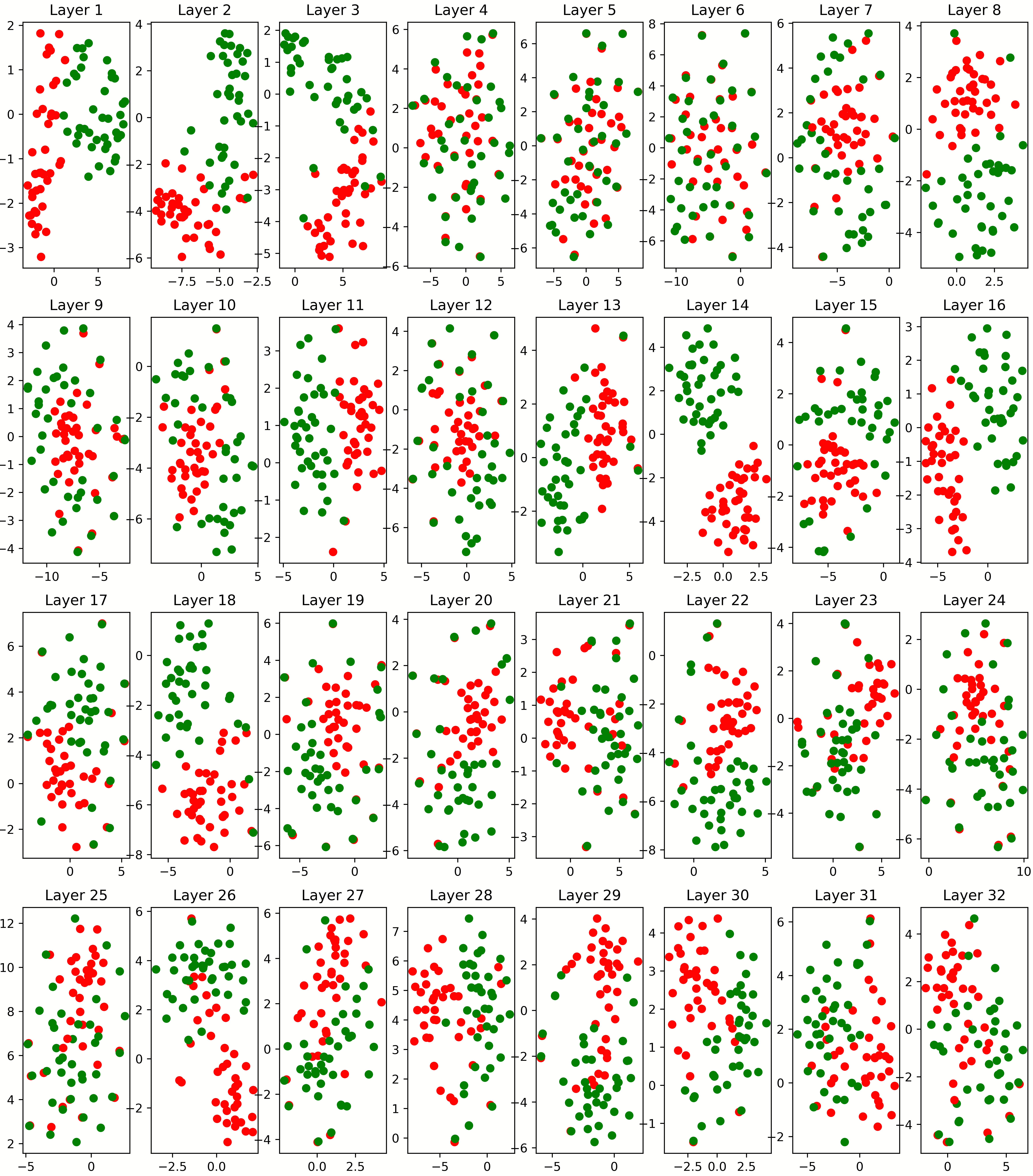} 
\caption{The visualization of t-SNE for each layer in the model Open-llama-7b. The red samples are MCQs, and the samples in green are LFGQs.}
\label{fig:67}
\end{center}
\end{figure}

\begin{figure}[H]
\begin{center}
\includegraphics[scale=0.03]{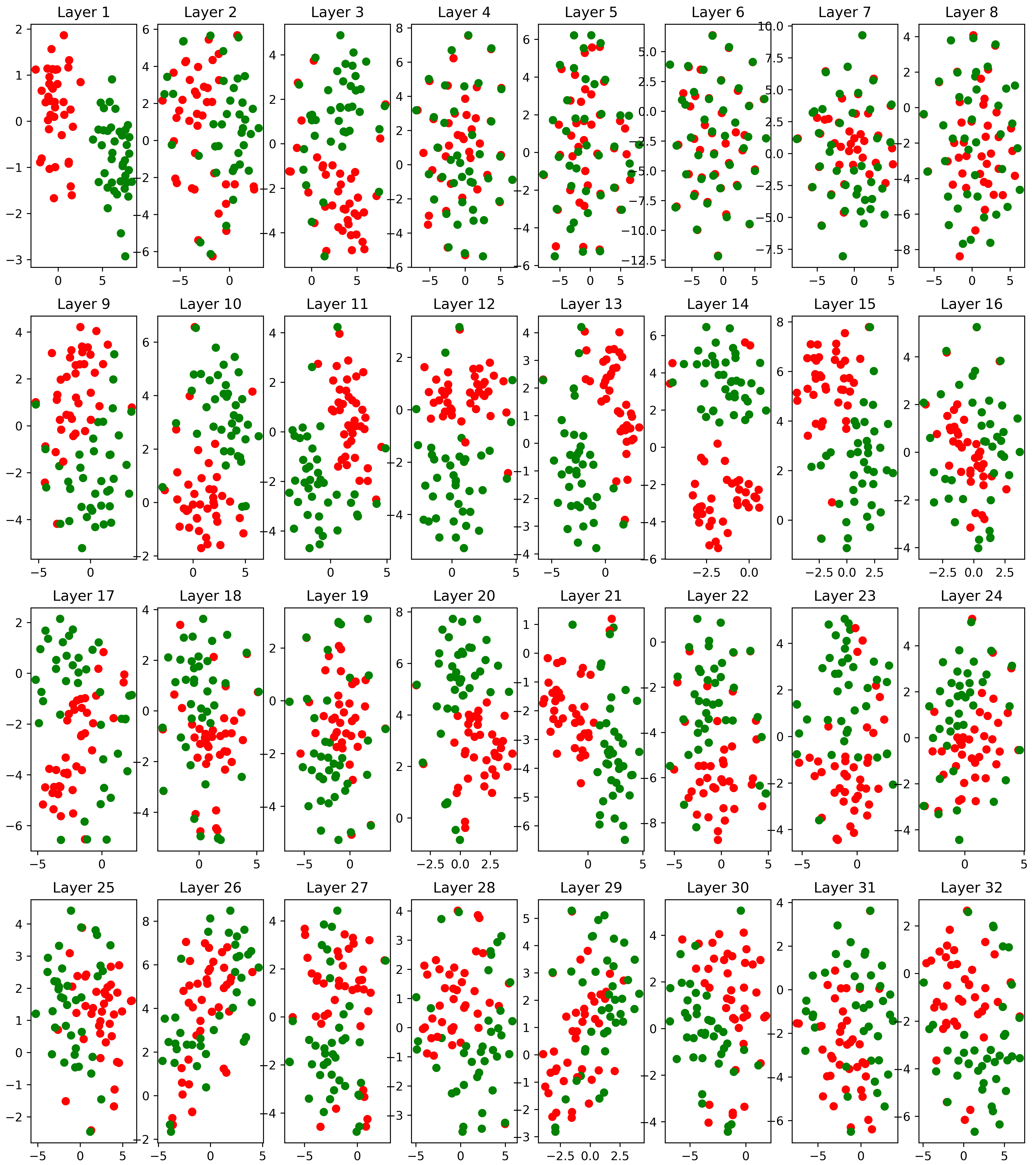} 
\caption{The visualization of t-SNE for each layer in the model Vicuna-7b-v1.3. The red samples are MCQs, and the samples in green are LFGQs.}
\label{fig:68}
\end{center}
\end{figure}

\begin{figure}[H]
\begin{center}
\includegraphics[scale=0.5]{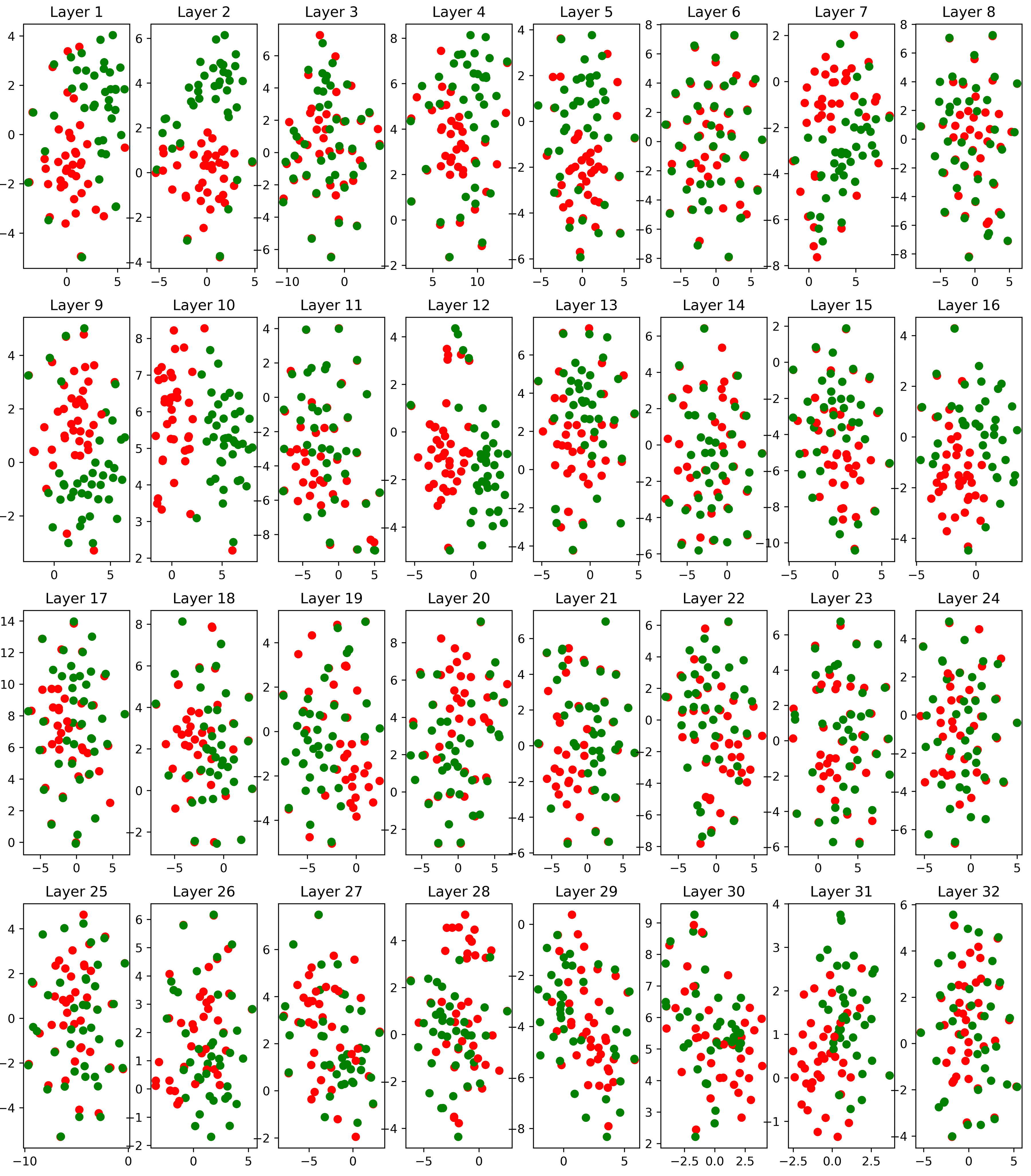} 
\caption{The visualization of t-SNE for each layer in the model Dolly-v2-3b. The red samples are MCQs, and the samples in green are LFGQs.}
\label{fig:69}
\end{center}
\end{figure}

\begin{figure}[H]
\begin{center}
\includegraphics[scale=0.03]{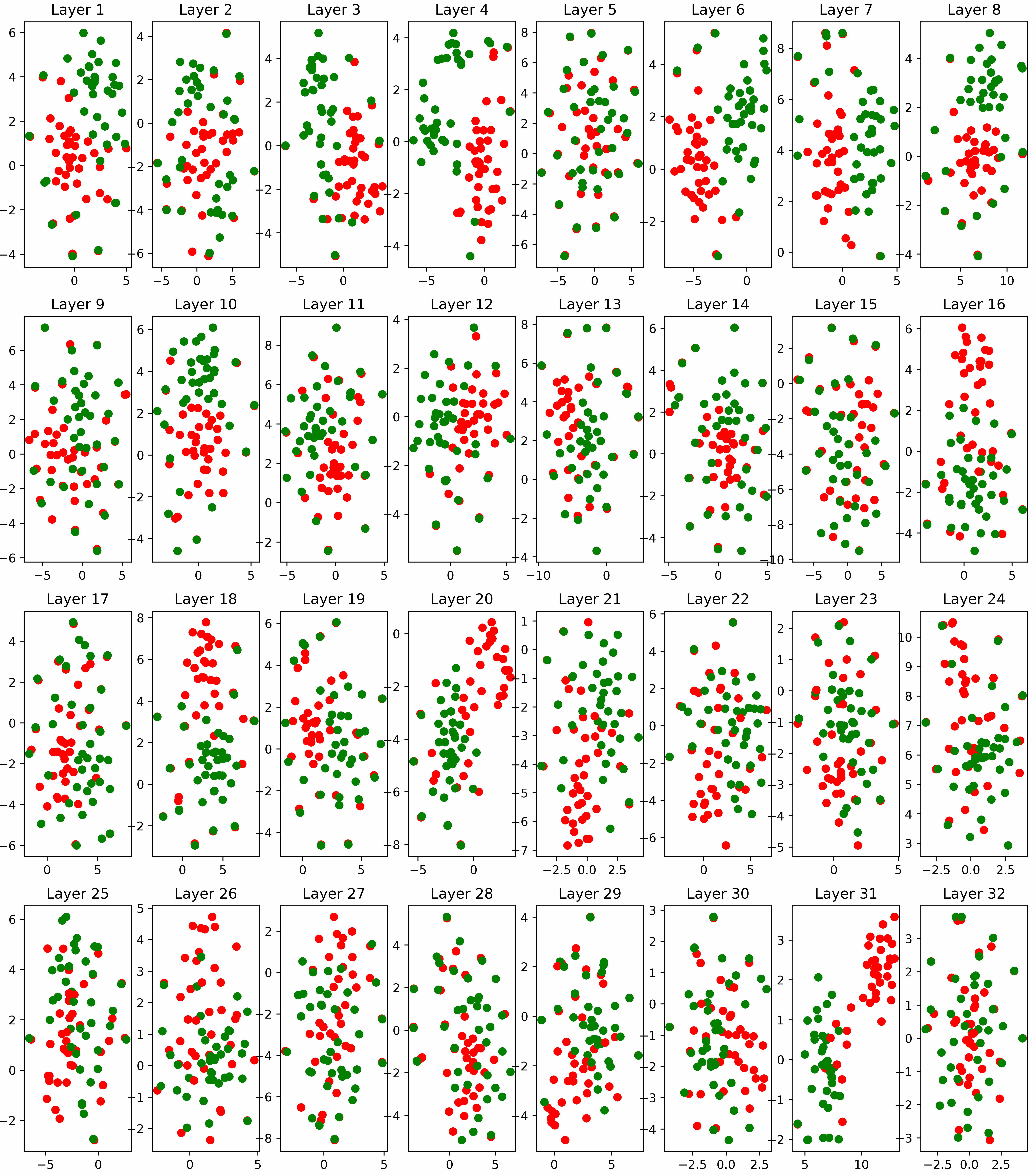} 
\caption{The visualization of t-SNE for each layer in the model Dolly-v2-7b. The red samples are MCQs, and the samples in green are LFGQs.}
\label{fig:70}
\end{center}
\end{figure}

\begin{figure}[H]
\begin{center}
\includegraphics[scale=0.03]{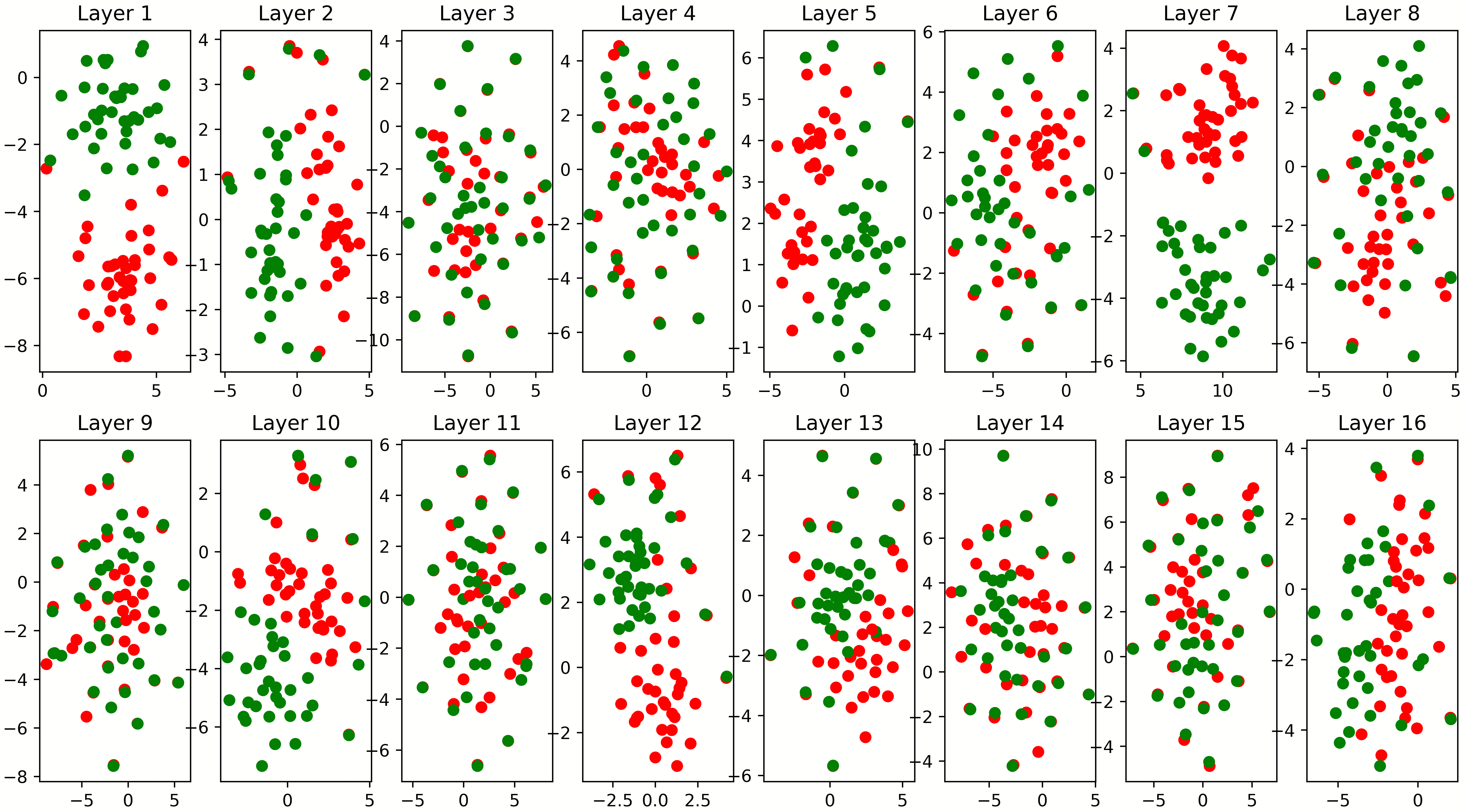} 
\caption{The visualization of t-SNE for each layer in the model Stablelm-tuned-alpha-3b. The red samples are MCQs, and the samples in green are LFGQs.}
\label{fig:72}
\end{center}
\end{figure}

\begin{figure}[H]
\begin{center}
\includegraphics[scale=0.03]{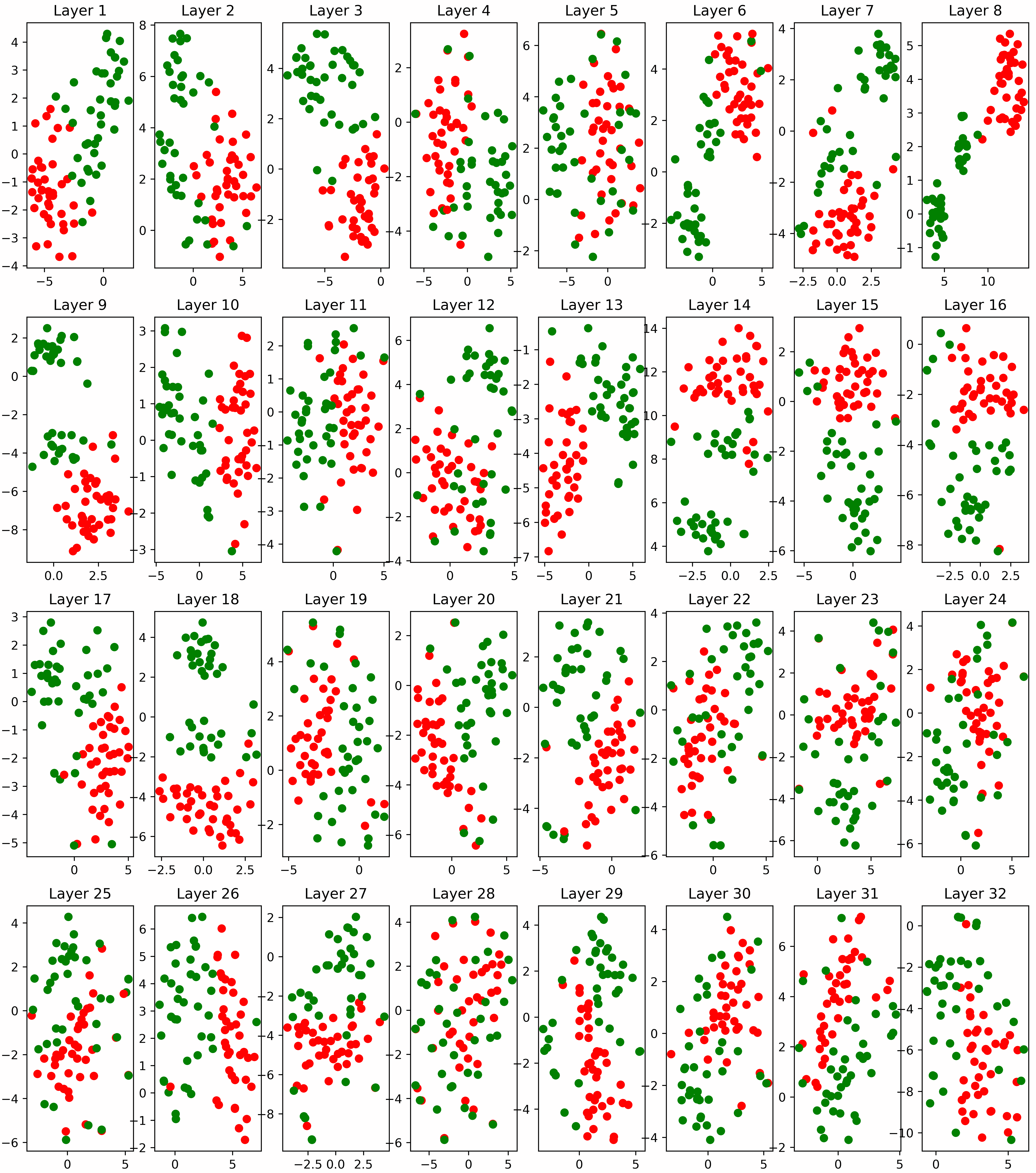} 
\caption{The visualization of t-SNE for each layer in the model RedPajama-INCITE-Instruct-3B. The red samples are MCQs, and the samples in green are LFGQs.}
\label{fig:71}
\end{center}
\end{figure}

\begin{figure}[H]
\begin{center}
\includegraphics[scale=0.03]{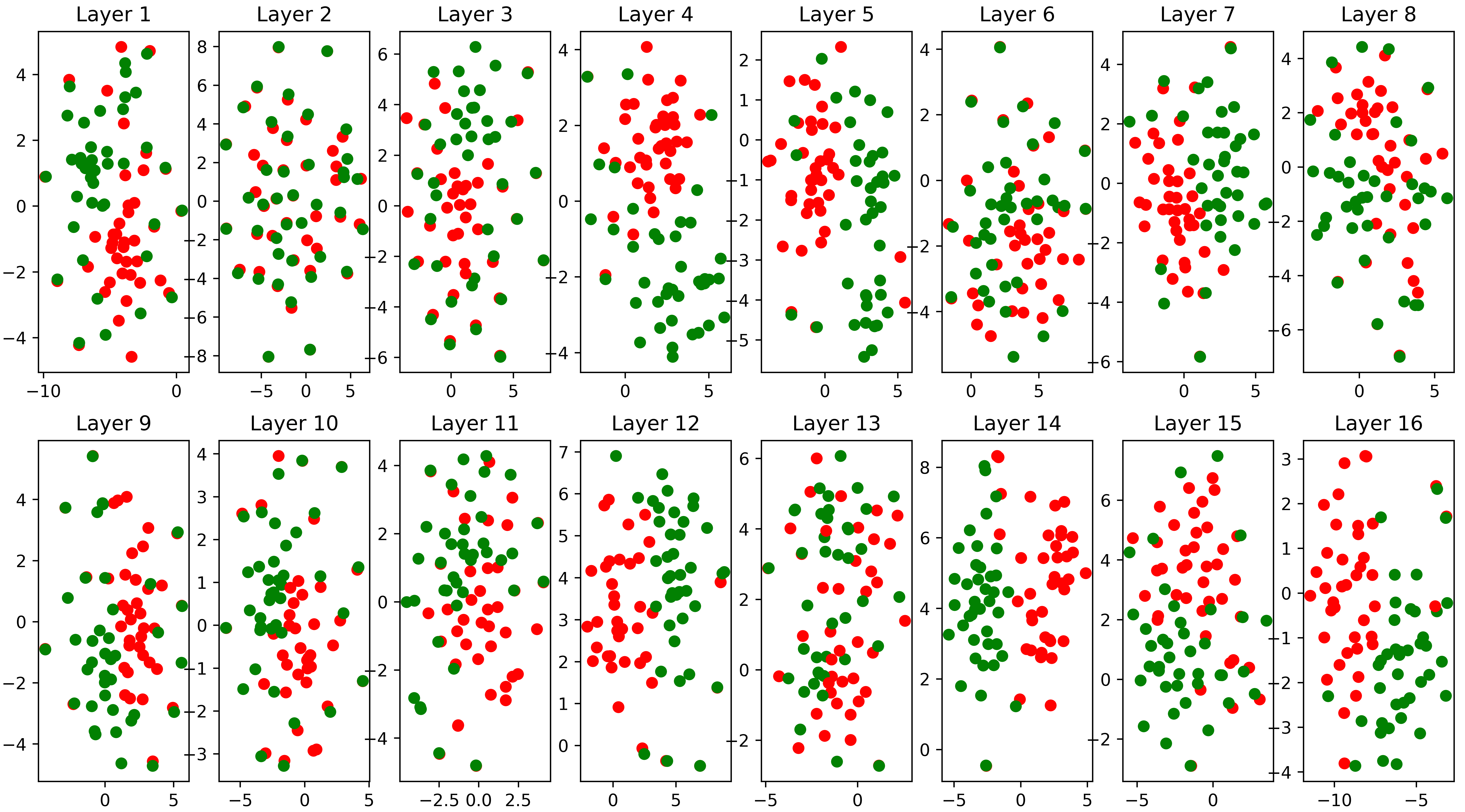} 
\caption{The visualization of t-SNE for each layer in the model Stablelm-tuned-alpha-7b. The red samples are MCQs, and the samples in green are LFGQs.}
\label{fig:73}
\end{center}
\end{figure}

In this appendix, we report the visualization of the t-SNE projected embeddings in the other seven models (Figures \ref{fig:66}-\ref{fig:73}) and the hidden embedding space details for each LLMs (Table \ref{tab:11}).

\begin{table*}[h]
\centering
\resizebox{0.98\textwidth}{!}{
\begin{tabular}{lccc}
\toprule
 \textbf{Model} & \textbf{hidden layers}  & \textbf{attention heads} &  \textbf{head dim.}  \\
\midrule
open-llama-3b & $26$ & $32$ & $100$   \\
open-llama-7b &$32$ & $32$ &$128$    \\
vicuna-7b-v1.3 &$32$ & $32$ &$128$   \\
dolly-v2-3b &$32$ & $32$ &$80$    \\
dolly-v2-7b &$32$ & $32$ &$128$   \\
Llama-2-7b-chat-hf &$32$ & $32$ &$128$    \\
RedPajama-INCITE-Instruct-3B-v1 &$32$ & $32$ &$80$   \\
stablelm-tuned-$\alpha$-3b & $16$ & $32$ &$128$   \\
stablelm-tuned-$\alpha$-7b & $16$ & $32$ &$192$    \\
\bottomrule
\end{tabular}}
\caption{Number of hidden layers, number of attention heads, and the head dimensionality of the LLMs.}
\label{tab:11}
\end{table*}

\end{document}